\documentclass[letterpaper]{article} 
\usepackage{aaai24}  
\usepackage{times}  
\usepackage{helvet}  
\usepackage{courier}  
\usepackage[hyphens]{url}  
\usepackage{graphicx} 
\usepackage{natbib}  
\usepackage{caption} 
\usepackage{algorithm}
\usepackage{algorithmic}

\usepackage{algorithm}
\usepackage{algorithmic}
\usepackage [switch]{lineno}
\usepackage{amsfonts}
\usepackage{amsbsy}
\usepackage{amsmath}
\usepackage{amssymb}
\usepackage{booktabs}
\usepackage{multirow}
\usepackage{epsfig}

\usepackage{newfloat}
\usepackage{listings}
\DeclareCaptionStyle{ruled}{labelfont=normalfont,labelsep=colon,strut=off} 
\floatstyle{ruled}
\newfloat{listing}{tb}{lst}{}
\floatname{listing}{Listing}
\title{Continuous Piecewise-Affine Based Motion Model for Image Animation}
\author{
    Hexiang Wang\textsuperscript{\rm 1$*$},
    Fengqi Liu\textsuperscript{\rm 1$*$},
    Qianyu Zhou\textsuperscript{\rm 1}\thanks{Equal Contribution.},
    Ran Yi\textsuperscript{\rm 1$\dagger$},
    Xin Tan\textsuperscript{\rm 2},
    Lizhuang Ma\textsuperscript{\rm 1,2}\thanks{Corresponding Authors.}
}
\affiliations{
    \textsuperscript{\rm 1} Department of Computer Science and Engineering, Shanghai Jiao Tong University, Shanghai, China \\
    \textsuperscript{\rm 2} East China Normal University, Shanghai, China\\
     \{whxsjtu123, liufengqi, zhouqianyu, ranyi\}@sjtu.edu.cn, xtan@cs.ecnu.edu.cn, ma-lz@cs.sjtu.edu.cn  \\
}

\usepackage{bibentry}

\begin{document}

\maketitle

\begin{abstract}

Image animation aims to bring static images to life according to driving videos and create engaging visual content that can be used for various purposes such as animation, entertainment, and education. Recent unsupervised methods utilize affine and thin-plate spline transformations based on keypoints to transfer the motion in driving frames to the source image. However, limited by the expressive power of the transformations used, these methods always produce poor results when the gap between the motion in the driving frame and the source image is large. To address this issue, we propose to model motion from the source image to the driving frame in highly-expressive diffeomorphism spaces. Firstly, we introduce Continuous Piecewise-Affine based (CPAB) transformation to model the motion and present a well-designed inference algorithm to generate CPAB transformation from control keypoints. Secondly, we propose a SAM-guided keypoint semantic loss to further constrain 
the keypoint extraction process and improve the semantic consistency between the corresponding keypoints on the source and driving images. 
Finally, we design a structure alignment loss to align the structure-related features extracted from driving and generated images, thus helping the generator generate results that are more consistent with the driving action. Extensive experiments on four datasets demonstrate the effectiveness of our method against state-of-the-art competitors quantitatively and qualitatively. Code will be publicly available at: https://github.com/DevilPG/AAAI2024-CPABMM.

\end{abstract}

\section{Introduction}

Given an input source image and a driving video, the image animation task aims to generate a video of the object in the source image performing the same action as the object in the driving video. It can be widely used in many real-world situations such as game production~\cite{Lele_Ross_Zhiyao_Chenliang_2019}, short video creation~\cite{Zhou_Sun_Wu_Loy_Wang_Liu_2021}, and film animation~\cite{Naruniec_Helminger_Schroers_Weber_2020}.

\begin{figure}[t]
    \centering
    \includegraphics[width=\linewidth]{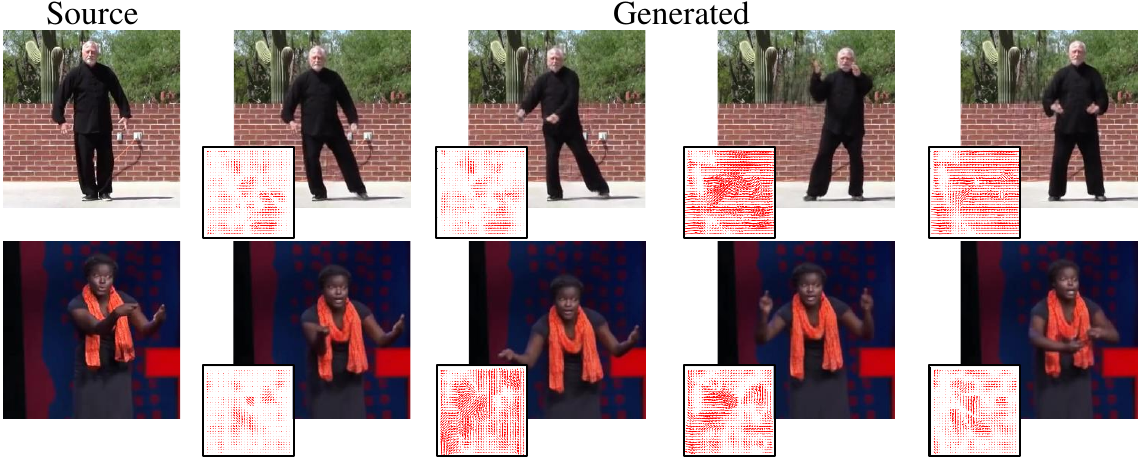}
    \caption{Our framework animates static image according to driving video via Continuous Piecewise-Affine velocity fields (inset), which divide the image space into a series of small grids (tessellation), and express the motion transfer from driving image to source image using independent affine transformations within each grid.} 
    \label{fig:1}
\end{figure}

Past researchers use prior knowledge and explicit structural information of specific objects (e.g. faces~\cite{wang2018video}, human bodies~\cite{ren2020human,ren2020deep}), such as keypoints, 3D models etc., to perform motion transfer, and achieve the goal of image animation with generation models such like GANs~\cite{goodfellow2020generative} and VAEs~\cite{kingma2013auto}. However, these methods usually rely on pre-trained models~\cite{cao2017realtime,booth2018large} to extract object-specific representations 
which are trained using expensive data annotations and cannot be used for arbitrary object categories.

Recent researchers~\cite{siarohin2019animating,siarohin2019first,siarohin2021motion,zhao2022thin,wang2022latent} design unsupervised end-to-end pipelines to perform motion transfer without prior knowledge about the object to be animated. 
Monkey-Net~\cite{siarohin2019animating} is the first to propose a framework with a keypoint detector, a dense motion network and a motion transfer generator, and trains the pipeline with reconstruction loss in an end-to-end manner. 
FOMM~\cite{siarohin2019first} and MRAA~\cite{siarohin2021motion} both use linear affine transformations to model motion, which are not expressive enough for this task.
TPSMM~\cite{zhao2022thin} utilizes more flexible non-linear thin-plate-spline (TPS) transformation to model motion and proposes to use multi-resolution occlusion masks in the generation module. However, TPS transformation tends to significantly distort the overall image to achieve keypoint alignment which can cause large destruction to the information in source image and thus be detrimental to preserving the identity of source object. Besides, their detected sets of keypoints are messy and the accuracy is not high, which limits the expressiveness of TPS transformation.

To address the above issues, we view the image animation task from two different perspectives:
transferring the motion of each driving frame to the source image, or transferring the appearance of the source image to each driving frame. These two different perspectives
correspond to two key challenges of this task: motion transfer and identity preservation. 
In order to achieve these two objectives, the flexibility and expressive power of the utilized transformations, as well as the accuracy of the detected keypoints, are crucial.

Based on these observations, we make several novel improvements to previous unsupervised methods. Firstly, we leverage Continuous Piecewise-Affine Based (CPAB) transformations~\cite{freifeld2017transformations} to model motion (see Figure~\ref{fig:1}), which is a simple yet highly expressive representation of complicated diffeomorphism spaces. Compared to TPS which performs image transformation in its entirely, CPAB applies transformation on the image in a piece-wise manner, which can better preserve the features of the original image in the course of motion transfer, thus significantly improving the ability to maintain the identity of object in the source image.
In this paper, we propose a carefully-designed gradient descent inference algorithm to estimate the corresponding CPAB transformation from a pair of keypoint sets. 
Secondly, we propose a Keypoint Semantic loss to further constrain the keypoint detector and improve the semantic consistency between the corresponding keypoints on source and driving images.
In order to obtain semantic information from the vicinity of keypoints on the image, we leverage the Segment Anything model~\cite{kirillov2023segment}
to extract keypoint-corresponded semantic information.

Finally, we extract structural information from generated and driving images with pre-trained DINO ViT model~\cite{caron2021emerging,tumanyan2022splicing} and design a Structure Alignment loss to assist the generator to generate results that are more consistent with the driving action. 

Our main contributions can be summarized as follows:

\begin{itemize}
    \item We introduce Continuous Piecewise-Affine Based (CPAB) transformation for image animation, which models motion from the source to the driving frame in highly-expressive diffeomorphism spaces. And we present an inference algorithm to generate CPAB transformation from control keypoints.
    \item  We first present a SAM-guided keypoint semantic loss to improve the results of keypoint detector. Besides, we design an innovative structure alignment loss to help the generator perform the motion transfer better. 
    \item  Extensive experiments and analysis demonstrate the effectiveness and superiority of our proposed method against state-of-the-art competitors on four datasets. 
\end{itemize}

\begin{figure*}[t]
    \centering
    \includegraphics[scale=0.48]{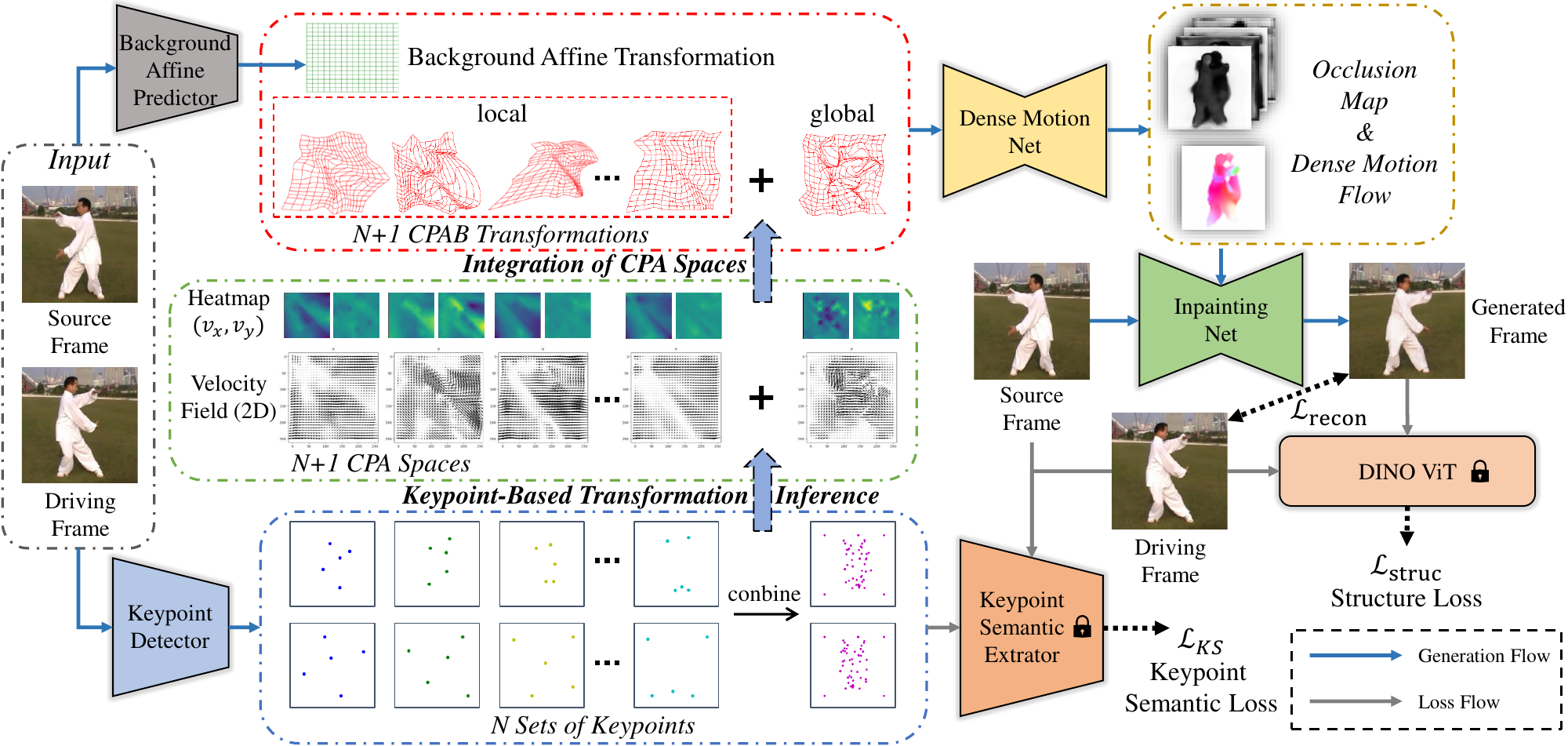}
    \caption{Overview of our method. The Keypoint Detector predicts $N$ sets of keypoints for source and driving images, from which we generate $N$ local CPA spaces with our Keypoint-Based Transformation Inference algorithm. Especially, we combine $N$ sets of keypoints and generate a global CPA space with all the keypoints. Then we integrate the $N+1$ CPA spaces and obtain $N+1$ CPAB transformations. Along with background transformation, the $N+2$ transformations are combined by the Dense Motion Net to generate several occlusion maps and dense optical flow. Finally, we feed the source image with dense motion results into Inpainting Net to generate the target frame. 
    } 
    \label{fig:framework}
\end{figure*}

\section{Related Work}

\subsection{Video Generation} Video generative methods have developed in diverse types, such as flow-based models~\cite{kingma2018glow}, variational autoencoders~\cite{kingma2013auto}, generative adversarial networks ~\cite{goodfellow2020generative}, and autoregressive models~\cite{chen2020generative}. 
With the rapid development of deep learning approaches, recent methods have started to introduce conditional variational autoencoders into the task of video generation. \cite{wang2018every} proposes a novel conditional multimodel network to tackle the one-to-many video generation task, which explicitly utilizes facial landmarks for synthesizing smile video sequences. By learning motion and content disentanglement in an unsupervised manner, MoCoGAN~\cite{tulyakov2018mocogan} presents a motion and content decomposed generative adversarial framework to map a sequence of random embeddings to a sequence of video frames to generate the final video. Our method utilizes a novel nonlinear warping formulation to generate the target video by predicting the dense motion between the source image and driving frames.

\subsection{Image Animation}
Previous methods~\cite{wang2018video,doukas2021headgan,ren2020deep} rely on explicit structural representations such as 2D landmarks and 3D models either annotated or extracted by pre-trained models to perform motion transfer on specific objects such as faces and human bodies. Recent unsupervised methods~\cite{siarohin2019first,siarohin2021motion,zhao2022thin} utilize video reconstruction task as self-supervision to train a framework in an end-to-end manner. 
FOMM~\cite{siarohin2019first} utilizes a first-order motion model which regresses affine transformation parameters from the source and driving keypoints to estimate dense motion and makes the generation network occlusion-aware by predicting occlusion maps in dense motion network. MRAA~\cite{siarohin2021motion} uses a PCA-based inference algorithm to calculate affine parameters which improves stability and performs better for articulated motions. It also proposes a background motion predictor to estimate background motion with an additional affine transformation. TPSMM~\cite{zhao2022thin} predicts several sets of keypoints from the source and driving image and uses each pair of keypoint sets to calculate a thin-plate-spline transformation to increase the flexibility of motion model, and proposes to predict multi-resolution occlusion maps. Compared with TPSMM, we use a gradient descent-based inference method to estimate Continuous Piecewise-Affine Based transformations and design a Keypoint Semantic loss based on SAM model~\cite{kirillov2023segment} to improve the keypoint detector.

\section{Method}
The overview of our method is shown in Figure~\ref{fig:framework}. The architecture of our framework is based on the TPSMM baseline~\cite{zhao2022thin}. The differences between TPSMM and our framework are as follows:
1) Instead of thin-plate-spline transformation, we leverage Continuous Piecewise-Affine Based (CPAB) transformation to estimate dense motion from source frame to driving frame. What's more, in order to fully take advantage of the expressive ability of CPAB, we further combine all sets of keypoints to generate a global CPAB transformation.  
2) In addition to equivariance loss, we further utilize Segment Anything Model~\cite{kirillov2023segment} to extract semantic information from the vicinity of keypoints on the image and design a Keypoint Semantic Loss to improve the performance of keypoint detector. 
3) We use pre-trained DINO ViT~\cite{caron2021emerging,tumanyan2022splicing} to extract structural-related features from the generated frame and driving frame and design a structure loss to generate better results.

\subsection{Preliminaries on CPAB Transformation}
\label{sec:3.1}
This section provides the definition and formulation of Continuous Piecewise-Affine Based transformation in 2D space. Let $\Omega$ be $\mathbb{R}^2$ or a certain subset of $\mathbb{R}^2$ (e.g. image space). Then a diffeomorphism $T:\Omega \rightarrow \Omega$ can be obtained via integration of some velocity fields. In our setting we choose spaces of $\Omega \rightarrow \mathbb{R}^2$ CPA velocity fields as foundation to construct CPAB transformation.

\begin{figure}[t]
    \centering
    \includegraphics[width=\linewidth]{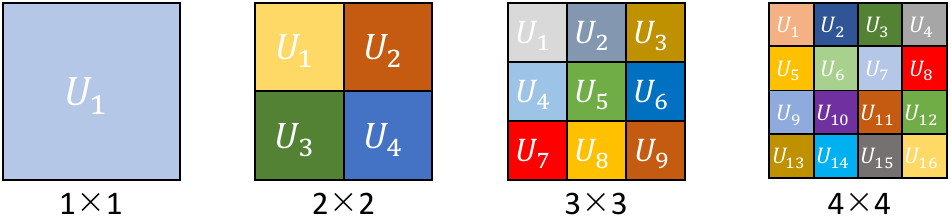}
    \caption{Visualization examples of type-II tessellation which divides image space into squares of same size.} 
    \label{fig:cpab}
\end{figure}

\noindent \textbf{Tessellation.}
The piecewise of CPAB is implemented as a certain type of tessellation, denoted by $\emph{P} = \{ U_c \}_{c=1}^{N_p}$, which is a set of $N_p$ closed subsets (cells) of $\Omega$ such that $U_1 \cup U_2 \cup \cdots \cup U_{N_p} = \Omega$ and the intersection of any adjacent cells is their shared border.  
We use type-II regular tessellation in \cite{freifeld2017transformations}, as shown in Figure~\ref{fig:cpab}.
We define an index function $Idx: \Omega \rightarrow \{ 1,\cdots,N_p \}$ which maps a point $x \in \Omega$ to the index of the cell $x$ is located.

\noindent \textbf{CPA velocity fields.}
Based on tessellation $\emph{P}$, a transformation $f: \Omega \rightarrow \mathbb{R}^2$ is defined as Piecewise-Affine (PA) w.r.t. $\emph{P}$ if $\{ f \, | \, U_c \}_{c=1}^{N_p}$ are all affine, i.e.: 
\begin{equation}
    f(x)=A_{Idx(x)} \Tilde{x},
\end{equation}
where $\Tilde{x}$ denotes the homogenous coordinates of $x$ and $A_c \in \mathbb{R}^{2\times3}$ is an affine transformation matrix. Based on this, if $f$ is continuous and PA, $f$ is called CPA. Denote $\Omega \rightarrow \mathbb{R}^2$ CPA velocity fields as $\emph{V}_{\Omega,\emph{P}}$ which depends on $\Omega$ and $\emph{P}$, and let $d=dim(\emph{V}_{\Omega,\emph{P}})$. Then for any given vector $\theta \in \mathbb{R}^d$ there exists a unique velocity field $v_{\theta}\in  \emph{V}_{\Omega,\emph{P}}$ that corresponds to $\theta$ one-to-one: $\theta \leftrightarrow v_{\theta}$,

where $\theta$ serves as control parameters of velocity field $v_{\theta}$.
Let $B=\{ B_1,\cdots,B_d \}$ denote a set of control basis calculated from constraint conditions, and let $A_{\theta} \triangleq (A_{1,\theta},\cdots,A_{N_p,\theta})$ denote the matrix parameters of piece-wise affine transformations corresponding to $v_{\theta}$, 
then according to derivation in \cite{freifeld2017transformations}, 
$A_{\theta}$ is calculated from $\theta$ and $B$ as follows:
\begin{equation}
\label{eqn:atheta}
    A_{\theta} = \text{vec}^{-1}(\sum_{i=1}^d \theta_i B_{i}),
\end{equation}
where vec denotes matrix vectorization operation.

\noindent \textbf{From CPA velocity fields to CPAB Transformation.}
Firstly for continuous $\Omega \rightarrow \mathbb{R}^2$ CPA velocity fields, we can define $\mathbb{R} \rightarrow \Omega$  trajectories for any $x\in \Omega$ and corresponding $v_{\theta} \in \emph{V}_{\Omega,\emph{P}}$ as $t \mapsto \Lambda_{\theta}(x,t)$. Here $t$ can be interpreted as movement time along trajectory, and $\Lambda_{\theta}(x,t)$ should be the solution of the following integral equation:
\begin{equation}
    \label{eqn:cpab_integ}
\Lambda_{\theta}(x,t) = x + \int_0^t v_{\theta}(\Lambda_{\theta}(x,k))dk \quad \text{where} \, \, v_{\theta}\in \emph{V}_{\Omega,\emph{P}}.
\end{equation}
The equivalent ordinary differential equation of Eq.~\ref{eqn:cpab_integ} is:
\begin{equation}
\label{eqn:cpab_integ1}
    \frac{d\Lambda_{\theta}(x,t)}{dt} = v_{\theta}(\Lambda_{\theta}(x,t)).
\end{equation}
Without loss of generality, fix $t=1$ and vary $x$, then $x \mapsto \Lambda_{\theta}(x,t)$ constitutes an $\Omega \rightarrow \Omega$ transformation, denoted as $T^{\theta}(x) \triangleq \Lambda_{\theta}(x,1) $.
Then with matrix exponential operation exp and condition $t=1$, the solution of Eq.~\ref{eqn:cpab_integ1} is $T^{\theta} = \text{exp}(v_{\theta})$. Based on this, the space of CPAB transformations $T_{\Omega,\emph{P}}$ is defined as:
\begin{equation}
\label{eqn:T}
    T_{\Omega,\emph{P}} \triangleq \text{exp}(\emph{V}_{\Omega,\emph{P}}) \triangleq \{ \text{exp}(v_{\theta}) \, | \, v_{\theta}\in  \emph{V}_{\Omega,\emph{P}} \}
\end{equation}

\subsection{CPAB Motion Model}
\label{sec:3.2}
As mentioned earlier, we utilize Continuous Piecewise-Affine Based (CPAB) transformations to estimate motion from the source image to the driving frame. Similar to \cite{zhao2022thin}, we use a Keypoint Detector to predict $N$ sets of keypoints, each set contains $5$ points. Our goal here is to obtain a fitting CPAB transformation $T_i^{\theta}$ for each pair of keypoints set denoted as $(kp_i^s,kp_i^d)$ where $ i \in [1,N] $ and $ kp_i^s,kp_i^d \in \mathbb{R}^{5\times 2}$. We divide this goal into two steps: 
we first obtain the corresponding CPA parameters $\theta_i$ based on each pair of keypoint sets $(kp_i^s,kp_i^d)$, and then calculate the specific transformation results through simulation integration of CPA spaces represented by $\theta_i$.

\noindent \textbf{Keypoint-Based Transformation Inference.} It has been clarified in Section~\ref{sec:3.1} that there is a one-to-one correspondence between $\theta$ and $v_{\theta}$ and we can obtain the affine transformations $A_{\theta}$ corresponding to $v_{\theta}$ with Eq.~\ref{eqn:atheta}. Thus the key here is to find an appropriate parameter $\theta_i$ such that $T_i^{\theta}$ can maximize alignment between $kp_i^s$ and $kp_i^d$ through warping. Here in our pipeline the warping function is back-warping, so the objective is $min\{ <T_i^{\theta}(kp_i^d) \, , kp_i^s>\}$ where $<\cdot \, ,\cdot>$ denotes some distance function. Considering the whole process of obtaining transformation results from given $\theta_i$ and keypoints pair $(kp_i^s,kp_i^d)$ (\textit{see details in Algorithm~1 in supplementary}) is differential, we can obtain the optimal $\theta_i$ with gradient descent. To be specific, we use a $\mathcal{L}_1$ loss $\mathcal{L}_{kp}$ to optimize $\theta_i$:
\begin{equation}
\label{eqn:lkp}
    \mathcal{L}_{kp} = E_i(|T_i^{\theta}(kp_i^d) - kp_i^s|_1).
\end{equation}
Through experiments, we have found that $200$ iterations of training are good enough for a fitting $\theta_i$.

\noindent \textbf{Integration of CPA Spaces.}
Given a CPA space with parameter $\theta \in \mathbb{R}^d$, tessellation $\emph{P}$, control basis $B$ and a set of points in $\mathbb{R}^2$, we can obtain the transformed points in a differential way. 
Using Eq.~(\ref{eqn:atheta}) we can obtain the affine matrix of each cell in tessellation $\emph{P}$. Then according to Eq.~(\ref{eqn:T}) we can calculate the CPAB transformation matrix of each cell. Based on this, for each point $x$ in the input grid we first find the cell number $c$ that it is located in with index function $Idx(x)$, then we use the corresponding transformation $T_{c,\theta}$ to transform $x$. After several steps of these operations, we obtain the final transformed results. The whole calculating process is summarized in Algorithm~1 in supplementary.
Note that we can use this algorithm to directly calculate $\mathcal{L}_{kp}$ in Eq.~(\ref{eqn:lkp}) by taking $kp_i^d$ as input. We can also generate optical flow of a certain $\theta$ by inputting a standard image grid of $\mathbb{R}^{H\times W \times 2}$ to Algorithm~1.

\noindent \textbf{Global CPAB Transformation Estimation.}
The CPAB transformation has strong expressive ability and is capable of performing keypoint-based transformation for a large number of keypoints,
which is a significant advantage over TPS transformation. To fully take advantage of CPAB transformation, we further combine all sets of keypoints and generate a global CPAB transformation $T_{N+1}^{\theta}$ which transforms all the keypoints, as shown in Figure~\ref{fig:framework}.

Following \cite{siarohin2021motion,zhao2022thin}, an extra background transformation $T^A$ is predicted by Background Affine Predictor. Then we use the total $N+2$ transformations to back-warp the source image. The $N+2$ warped images are cascaded and fed into Dense Motion Net ~\cite{newell2016stacked} to predict $N+2$ confidence maps $M_k \in \mathbb{R}^{H\times W}, k \in [0,N+1]$, which are then used to combine all the transformations and generate the final dense optical flow $T$ as:
\begin{equation}
    T(p) = M_0(p)T^A(p) + \sum_{k=1}^{N+1} M_k(p)T_k^{\theta}(p),
\end{equation}
where $p \in \Omega$ denotes pixel coordinates in image space. The dense optical flow $T$ is then fed into following Inpainting Net to warp the feature maps of the source frame and reconstruct the driving frame (Figure~\ref{fig:framework}).

\subsection{SAM-Guided Keypoint Semantic Loss}
\label{sec:3.3}
The transformation estimation in this unsupervised end-to-end framework is based on keypoints, however, without a direct constraint on the keypoints, the predicted keypoints
are messy, {\it i.e.,} the $i$-th keypoint in source image has different semantics with the $i$-th keypoint in driving image. 
We aim to improve the semantic consistency between corresponding keypoints,
{\it i.e.,} the keypoint in the source image should be in the same part of the object as the corresponding keypoint in the driving frame. This requirement is very challenging 
because all the keypoints are predicted without supervision. Based on this observation, 
we propose to extract semantic information from the vicinity of keypoints on the image with pre-trained SAM, and design a Keypoint Semantic loss which directly constrains the one-to-one correspondence between source and driving keypoints.

\begin{figure}[t]
    \centering
    \includegraphics[width=\linewidth]{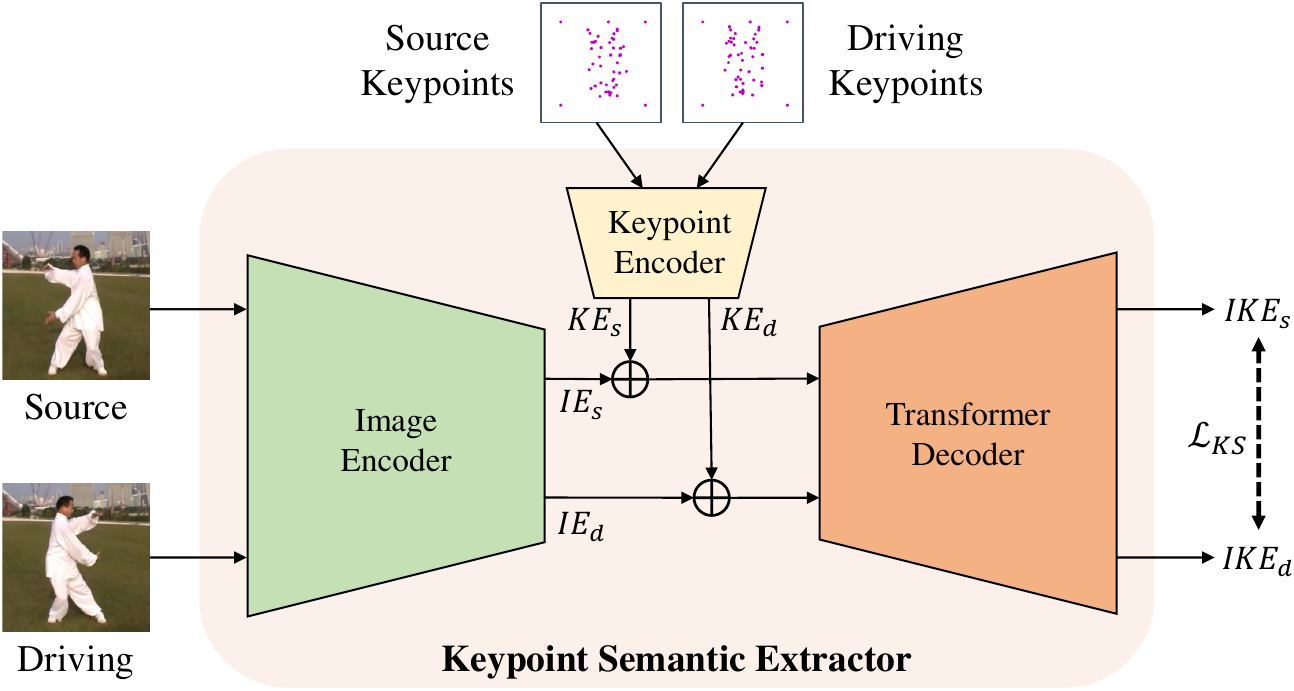}
    \caption{The structure of Keypoint Semantic Extractor. The image encoder, keypoint encoder and transformer decoder are extracted from pre-trained SAM~\cite{kirillov2023segment}.} 
    \label{fig:sam}
\end{figure}

As shown in Figure~\ref{fig:sam}, we build a Keypoint Semantic Extractor using the ViT image encoder, the keypoint encoder, and the transformer decoder from pretrained SAM. The image encoder extracts image embeddings $IE_s$ and $IE_d$ from the source and driving image, and the keypoint encoder extracts keypoint embeddings $KE_s$ and $KE_d$ from source and driving keypoints separately. The transformer decoder takes image embedding $(IE_s, IE_d)$ and keypoint embedding $(KE_s, KE_d)$ as inputs, and outputs Image-Keypoint Embeddings ($IKE_s, IKE_d$), which contain the semantic information of the vicinity of the keypoints on the image.
We propose the Keypoint Semantic loss $\mathcal{L}_{KS}$ to maximize the cosine similarity between $IKE_s$ and $IKE_d$:
\begin{equation}
    \mathcal{L}_{KS} = \sum_{i=1}^{N\times 5}(1-\frac{IKE_{s,i}\cdot IKE_{d,i}}{\max (|| IKE_{s,i}||\cdot ||IKE_{d,i} ||,\epsilon)}),
\end{equation}
where $IKE_{s,i}, IKE_{d,i}$ denote the semantic image-keypoint embeddings of $i$-th keypoint in source and driving image.

\subsection{Structural Alignment Loss From DINO ViT}
\label{sec:3.4}

Our training process is self-supervised as an image reconstruction task. The whole framework is trained to generate images that have the same structure as the driving frame while maintaining the appearance of the source frame. 
In our training process, the appearance information of source and driving images are always the same, so the key here is to transfer motion from source to driving, which can also be interpreted as generating images with a structure consistent with driving frames. 
A recent work~\cite{tumanyan2022splicing} found the global CLS token from deep layers of ViT provides a powerful representation of visual appearance, while the self-similarity of keys from last layer of ViT captures structural information.

To improve the structure consistency between driving frame and generated image, we feed 
them into pre-trained DINO ViT~\cite{caron2021emerging} separately and calculate the self-similarity matrices of keys in last layer, denoted as $SK_d$ and $SK_g$, then the Structure Alignment loss is calculated as:
\begin{equation}
    \mathcal{L}_{struc} = || SK_g - SK_d ||_F.
\end{equation}

\subsection{Training Losses}
We use the VGG-19 perceptual loss~\cite{johnson2016perceptual} between driving frame $I_d$ 
and generated frame $I_g$ as a reconstruction loss:
\begin{equation}
    \mathcal{L}_{perc} = \sum_j \sum_i \lvert V_i(I_d^{(j)}) -V_i(I_g^{(j)}) \rvert,
\end{equation}
where $V_i$ denotes the $i^{th}$ layer of pre-trained VGG-19 network, $I_d^{(j)}$ and $I_g^{(j)}$ denote driving frame and generated frame downsampled for $j$ times.
We further follow TPSMM to use a warp loss as:
\begin{equation}
    \mathcal{L}_{warp} = \sum_i \lvert T(E_i(I_s))-E_i(I_d) \rvert,
\end{equation}
where $I_s$ is source image and $E_i$ denotes the $i^{th}$ layer of encoder in Inpainting Net. The total reconstruction loss is:
\begin{equation}
    \mathcal{L}_{recon} = \mathcal{L}_{perc} + \mathcal{L}_{warp}.
\end{equation}

For regular constraint of Keypoint Detector, we also use equivariance loss:
\begin{equation}
    \mathcal{L}_{eq} = \lvert E_{kp}(T^{rand}(I_d))-T^{rand}(E_{kp}(I_d)) \rvert,
\end{equation}
where $E_{kp}$ is Keypoint Detector and $T^{rand}$ denotes a random TPS transformation.

Overall, the total training loss function is:
\begin{equation}
    \mathcal{L} = \lambda_{r}\mathcal{L}_{recon} + \lambda_{e}\mathcal{L}_{eq} + \lambda_{k}\mathcal{L}_{KS} + \lambda_{s}\mathcal{L}_{struc},
\end{equation}
where $\lambda_{r},\lambda_{e},\lambda_{k},\lambda_{s}$ are hyper-parameters.

\begin{table*}[t]
    \centering
    \begin{tabular}{c|ccc|ccc|ccc|c}
    \toprule
         & \multicolumn{3}{c|}{TaiChiHD} & \multicolumn{3}{c|}{TED-talks} & \multicolumn{3}{c|}{VoxCeleb} & MGif \\
         & $\mathcal{L}_1$ & (AKD, MKR) & AED & $\mathcal{L}_1$ & (AKD, MKR) & AED &
         $\mathcal{L}_1$ & AKD & AED & $\mathcal{L}_1$  \\
    \midrule
    X2Face & 0.080 & (17.65, 0.109) & 0.27 & - & - & - & 0.078 & 7.69 & 0.405 & - \\
    Monkey-Net & 0.077 & (10.80, 0.059) & 0.288 & - & - & - & 0.049 & 1.89 & 0.199 & - \\
    FOMM & 0.055 & (6.62, 0.031) & 0.164 & 0.033 & (7.07, 0.014) & 0.163 & 0.041 & 1.29 & 0.135 & 0.0225 \\
    MRAA & 0.048 & (5.41, 0.025) & 0.149 & 0.026 & (4.01, 0.012) & 0.116 & 0.040 & 1.29 & 0.136 & 0.0274 \\
    TPSMM & 0.045 &  (\textbf{4.57}, \textbf{0.018}) & 0.151 & 0.027 & (3.39, \textbf{0.007}) & 0.124 & 0.039 & \textbf{1.22} & 0.125 & 0.0212 \\
    Ours & \textbf{0.041} & (4.61, 0.021) & \textbf{0.117} & \textbf{0.022} & (\textbf{3.21}, 0.008) & \textbf{0.085} & \textbf{0.036} & 1.25 & \textbf{0.121} & \textbf{0.0169} \\
    \bottomrule
    \end{tabular}
    \normalsize
    \caption{Quantitative comparison of video reconstruction task with the state of the art on four different datasets. 
        }
    \label{tab:qcomp}
\end{table*}

\section{Experiments}

\subsection{Datasets}
Following TPSMM~\cite{zhao2022thin}, we train and test our method on four datasets of different object categories (e.g. human bodies, faces, pixel animals). We use the same data pre-processing protocol and train-test split strategy as in \cite{siarohin2021motion}. The datasets are as follows:
\begin{itemize}
    \item \emph{TaiChiHD}~\cite{siarohin2019first} consists of videos showcasing full-body TaiChi performances downloaded from YouTube and cropped to a resolution of $256\times 256$ based on the bounding boxes around the performers.
    \item \emph{TED-talks}~\cite{siarohin2021motion} contains videos of TED talks downloaded from YouTube, which were downscaled to $384 \times 384$ resolution based on the upper human bodies. The video length ranges from $64$ to $1024$ frames.
    \item \emph{VoxCeleb}~\cite{nagrani2017voxceleb} comprises of videos featuring various celebrities speaking, which were downloaded from YouTube and cropped to a resolution of $256\times 256$ based on the bounding boxes surrounding their faces. Video length ranges from $64$ to $1024$ frames.
    \item \emph{MGif}~\cite{siarohin2019animating} is a collection of .\emph{gif} files featuring pixel animations of animals in motion, which was obtained through Google searches.
\end{itemize}

\subsection{Experiment Settings}
\textbf{Evaluating Protocols.}
There are two different sets of performance evaluating protocols: 1) Image reconstruction, where the first frame of each video is used as source image and subsequent frames are used as driving frames for animation, making driving frames the ground truth, allowing for quantitative evaluation. 
2) Image animation, which represents the actual
usage scenario, a single image is paired with a driving video of a different object for animation. In this case, there is no ground truth, making it impossible to evaluate with quantitative metrics. Typically, performance is evaluated by visual comparison and user study.

\noindent \textbf{Evaluation Metrics.}
Strictly following previous video reconstruction protocols, we use the commonly-used metrics for evaluation,  including $\mathcal{L}_1$, AKD (average keypoint distance), MKR (missing keypoint rate) and AED (average euclidean distance). Please refer to the supplementary material for the specific definitions of these metrics.

\noindent \textbf{Implementation Details.}
We set $N=10$ to use $10$ local CPAB transformations and $1$ global CPAB transformation. 
We implement the framework with PyTorch and use Adam optimizer to update our model. We used one GeForce RTX 3090 GPU to train our model for $100$ epochs in all datasets with an initial learning rate of $0.0001$. VoxCeleb, TaiChiHD, and MGif for seven days, while TED-talks was trained for fifteen days. 
We set the training hyper-parameters as: $\lambda_{r}=10,\lambda_{e}=10,\lambda_{k}=1,\lambda_{s}=0.1$. 

\begin{figure}[t]
    \centering
    \includegraphics[width = 0.9\linewidth]{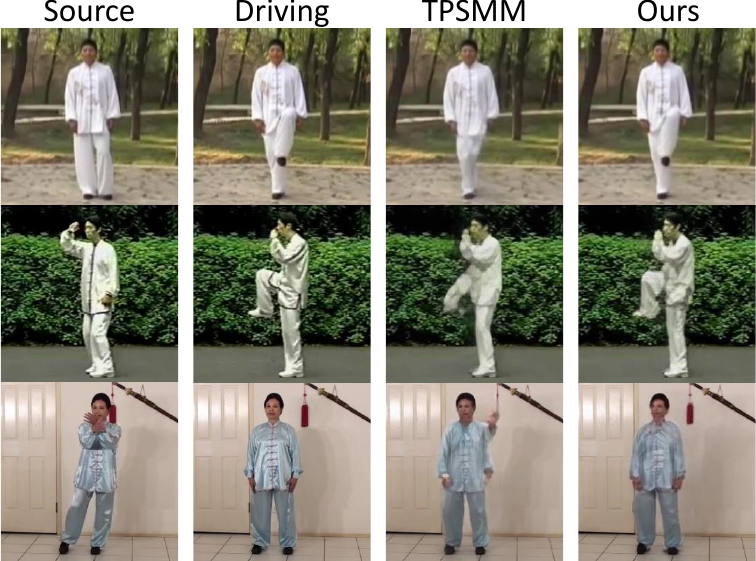}
    \caption{Video reconstruction task: some bad cases generated by TPSMM, while our method performs better.} 
    \label{fig:image_recon_comp}
\end{figure}

\begin{figure*}[t]
    \centering
    \includegraphics[scale=0.5]{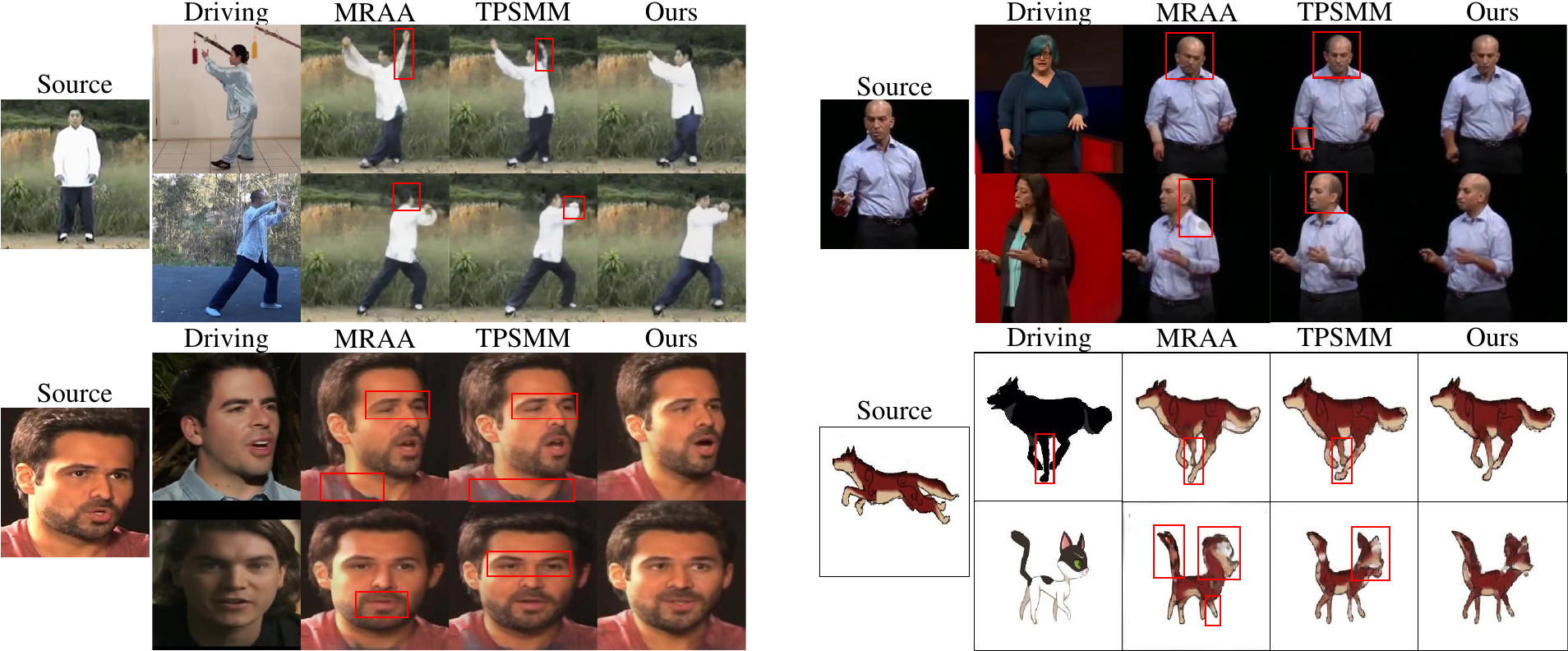}
    \caption{Qualitative comparison with MRAA and TPSMM on image animation task: TaiChiHD (top left), TED-talks (top right), VoxCeleb (bottom left), MGif (bottom right). \textit{Please refer to the supplementary material for more examples.}}
    \label{fig:maincomp}
\end{figure*}

\subsection{Quantitative Evaluation}
We quantitatively compare our method with previous baseline methods: X2Face~\cite{wiles2018x2face}, Monkey-Net~\cite{siarohin2019animating}, FOMM~\cite{siarohin2019first}, MRAA~\cite{siarohin2021motion} and TPSMM~\cite{zhao2022thin}. The quantitative results of video reconstruction task on four datasets are shown in Table~\ref{tab:qcomp}. Our method achieves significant improvements on $\mathcal{L}_1$ and AED metrics, which indicates that our model can reconstruct images closer to real ones and better preserve object identity. However, our AKD and MKR metrics on TaiChiHD and VoxCeleb datasets are slightly worse than TPSMM. This might be caused by the discrepant properties between TPS and CPAB transformations. TPS transformation prioritizes aligning keypoints and tends to significantly distort the overall image to achieve keypoint alignment, thus enabling more direct motion transfer. In contrast, CPAB transformation tends to adjust the interior of the image using different velocity fields in each tessellation cell to perform warping while maintaining the overall image shape, thus better preserving overall information. 

\begin{table}[t]
    \centering
    \begin{tabular}{c|ccc}
    \toprule
         & $\mathcal{L}_1$ & (AKD,MKR) & AED \\
        \midrule
       TPSMM &  0.045 & \textbf{(4.57,0.018)} &  0.151 \\
       +CPAB & 0.044 & (5.10,0.024) & 0.123 \\
       +CPAB,Global & 0.044 & (4.96,0.023) & 0.121 \\
       +CPAB,Global,$\mathcal{L}_{KS}$ & 0.043 & (4.78,0.024) & 0.119 \\
       +CPAB,Global,$\mathcal{L}_{struc}$ & 0.042 & (4.81,0.023) & 0.122 \\
       full model & \textbf{0.041} & (4.61,0.021) & \textbf{0.117} \\
       \bottomrule
    \end{tabular}
    \caption{Ablation of key components of proposed method. CPAB means using CPAB transformation to model motion. Global means adding an extra global CPAB transformation.}
    \label{tab:ablation}
\end{table}

\begin{figure}[t]
    \centering
    \includegraphics[width = \linewidth]{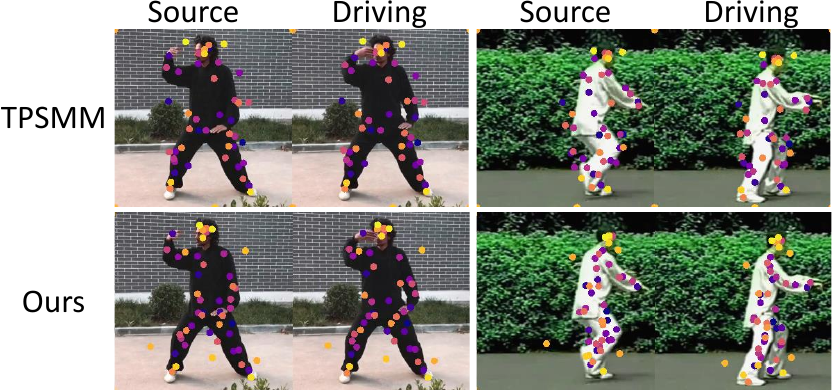}
    \caption{Examples of detected keypoints. Our method detects more keypoints that are evenly distributed across the object with higher quality.
    Yellow points on the background detected by our method capture the background motion.} 
    \label{fig:kp_comp}
\end{figure}


\subsection{Qualitative Evaluation}
Here we qualitatively compare our method with the current state-of-the-art method
TPSMM with two protocols: video reconstruction and image animation, and conduct a user study to subjectively compare image animation results.

\noindent \textbf{Video Reconstruction.} 
Figure~\ref{fig:image_recon_comp} shows video reconstruction results of several challenging cases from TaiChiHD dataset. In the first row, the motion discrepancy between source and driving images is minor but ambiguous. TPSMM fails to reconstruct the left leg while our method succeeds. In the second row, the human lifts his left hand and right leg at the same time. TPSMM can only generate a lifted left hand and misses the lifted right leg, while our method can reconstruct both actions. In the third row, the human drops her hands and stands at a new location to the right of the original position. For TPSMM, there exists a breakup hand part that has not been put down and is obscured with the background. Our method clearly reconstructs the image with higher quality. What's more, in Figure~\ref{fig:kp_comp} we show some examples of detected keypoints. With the keypoint semantic loss, keypoints detected by our keypoint detector are evenly distributed on the object with higher accuracy. 

\noindent \textbf{Image Animation.} 
Some image animation results on four datasets comparing with TPSMM and MRAA are shown in Figure~\ref{fig:maincomp}. For fair comparison, we use the \emph{standard} mode to generate animation results for all methods. The results demonstrate that our method outperforms MRAA and TPSMM in terms of overall quality of generated image, maintenance of identity and image details, while comparable to TPSMM in terms of motion transfer.

\noindent \textbf{User Study.} 
We conducted a subjective user study to compare our method with MRAA and TPSMM. We randomly selected 20 videos generated by MRAA, TPSMM and our method on four datasets (five videos per dataset) and invited $35$ volunteers to watch them, selecting the ones they preferred the most. We calculated proportion of votes obtained by each method on each dataset, and our method received more than half of the votes in all four datasets with proportions of $56.57\%,50.86\%,55.43\%$ and $65.71\%$ separately. 

\subsection{Ablation Study}
In this section, we conduct comprehensive ablation experiments on TaiChiHD dataset to systematically evaluate the contribution of key components to the final performance of the proposed method. We add our proposed components in turn and compare them with MRAA and TPSMM on video reconstruction metrics. The results are shown in Table~\ref{tab:ablation}. 

In the third row of Table~\ref{tab:ablation} we replace the $10$ TPS transformations with $10$ CPAB transformations with tessellation size of $4\times 4$, resulting in significant improvement on AED metric, demonstrating CPAB transformation's advantage of preserving identity. In the fourth row of Table~\ref{tab:ablation} we add an extra global CPAB transformation with a tessellation size of $6 \times 6$ which improves AKD and MKR, indicating better motion transfer. In the fifth row of Table~\ref{tab:ablation}, we add our Keypoint Semantic loss in the training process, which achieves improvement in all metrics, proving the importance of the quality of keypoint prediction for exerting the expression ability of transformation. However, after adding $\mathcal{L}_{KS}$ there exists a slightly higher MKR, which is not expected.
In the sixth row, we add our Structure Alignment loss in the training process, which successfully improves $\mathcal{L}_1$, AKD and MKR.

\section{Conclusion}
In this paper, we propose an unsupervised image animation method leveraging continuous piecewise-affine velocity field to estimate motion from the source image and generate corresponding optical flow. We utilize Segment Anything model to extract semantic information for keypoints and design a Keypoint Alignment loss to improve the performance of keypoint detector. Then, we use pre-trained DINO ViT to extract the structural-related information between the generated image and driving image and utilize a Structural Alignment loss to improve the motion transfer ability of the inpainting net. Experiments demonstrate that our method achieves state-of-the-art performance on most benchmarks.

\section{Acknowledgments}
This work is sponsored by National Natural Science Foundation of China (No. 72192821, 62302297, 62302167), Shanghai Municipal Science and Technology Major Project (2021SHZDZX0102), Shanghai Science and Technology Commission (21511101200), Shanghai Sailing Program (22YF1420300, 23YF1410500), CCF-Tencent Open Research Fund (RAGR20220121), Young Elite Scientists Sponsorship Program by CAST (2022QNRC001), CCF-Tencent Rhino-Bird Young Faculty Open Research Fund (RAGR20230121).
This work was also supported by Shenzhen SmartMore Corporation, Ltd, which provided GPUs and computing resources for us. 

\bibliography{aaai24}

\begin{thebibliography}{28}
\providecommand{\natexlab}[1]{#1}

\bibitem[{Booth et~al.(2018)Booth, Roussos, Ponniah, Dunaway, and Zafeiriou}]{booth2018large}
Booth, J.; Roussos, A.; Ponniah, A.; Dunaway, D.; and Zafeiriou, S. 2018.
\newblock Large scale 3d morphable models.
\newblock \emph{International Journal of Computer Vision}, 126(2): 233--254.

\bibitem[{Cao et~al.(2017)Cao, Simon, Wei, and Sheikh}]{cao2017realtime}
Cao, Z.; Simon, T.; Wei, S.-E.; and Sheikh, Y. 2017.
\newblock Realtime multi-person 2d pose estimation using part affinity fields.
\newblock In \emph{Proceedings of the IEEE Conference on Computer Vision and Pattern Recognition}, 7291--7299.

\bibitem[{Caron et~al.(2021)Caron, Touvron, Misra, J{\'e}gou, Mairal, Bojanowski, and Joulin}]{caron2021emerging}
Caron, M.; Touvron, H.; Misra, I.; J{\'e}gou, H.; Mairal, J.; Bojanowski, P.; and Joulin, A. 2021.
\newblock Emerging properties in self-supervised vision transformers.
\newblock In \emph{Proceedings of the IEEE/CVF International Conference on Computer Vision}, 9650--9660.

\bibitem[{Chen et~al.(2020)Chen, Radford, Child, Wu, Jun, Luan, and Sutskever}]{chen2020generative}
Chen, M.; Radford, A.; Child, R.; Wu, J.; Jun, H.; Luan, D.; and Sutskever, I. 2020.
\newblock Generative pretraining from pixels.
\newblock In \emph{International Conference on Machine Learning}, 1691--1703. PMLR.

\bibitem[{Doukas, Zafeiriou, and Sharmanska(2021)}]{doukas2021headgan}
Doukas, M.~C.; Zafeiriou, S.; and Sharmanska, V. 2021.
\newblock Headgan: One-shot neural head synthesis and editing.
\newblock In \emph{Proceedings of the IEEE/CVF International Conference on Computer Vision}, 14398--14407.

\bibitem[{Freifeld et~al.(2017)Freifeld, Hauberg, Batmanghelich, and Fisher}]{freifeld2017transformations}
Freifeld, O.; Hauberg, S.; Batmanghelich, K.; and Fisher, J.~W. 2017.
\newblock Transformations based on continuous piecewise-affine velocity fields.
\newblock \emph{IEEE Transactions on Pattern Analysis and Machine Intelligence}, 39(12): 2496--2509.

\bibitem[{Goodfellow et~al.(2020)Goodfellow, Pouget-Abadie, Mirza, Xu, Warde-Farley, Ozair, Courville, and Bengio}]{goodfellow2020generative}
Goodfellow, I.; Pouget-Abadie, J.; Mirza, M.; Xu, B.; Warde-Farley, D.; Ozair, S.; Courville, A.; and Bengio, Y. 2020.
\newblock Generative adversarial networks.
\newblock \emph{Communications of the ACM}, 63(11): 139--144.

\bibitem[{Johnson et~al.(2016)Johnson, Alahi, Fei-Fei et~al.}]{johnson2016perceptual}
Johnson, J.; Alahi, A.; Fei-Fei, L.; et~al. 2016.
\newblock Perceptual losses for real-time style transfer and super-resolution.
\newblock In \emph{Computer Vision--ECCV 2016: 14th European Conference, Amsterdam, The Netherlands, October 11-14, 2016, Proceedings, Part II 14}, 694--711. Springer.

\bibitem[{Kingma and Dhariwal(2018)}]{kingma2018glow}
Kingma, D.~P.; and Dhariwal, P. 2018.
\newblock Glow: Generative flow with invertible 1x1 convolutions.
\newblock \emph{Advances in Neural Information Processing Systems}, 31.

\bibitem[{Kingma et~al.(2013)}]{kingma2013auto}
Kingma, D.~P.; et~al. 2013.
\newblock Auto-encoding variational bayes.
\newblock \emph{arXiv preprint arXiv:1312.6114}.

\bibitem[{Kirillov et~al.(2023)Kirillov, Mintun, Ravi, Mao, Rolland, Gustafson, Xiao, Whitehead, Berg, Lo et~al.}]{kirillov2023segment}
Kirillov, A.; Mintun, E.; Ravi, N.; Mao, H.; Rolland, C.; Gustafson, L.; Xiao, T.; Whitehead, S.; Berg, A.~C.; Lo, W.-Y.; et~al. 2023.
\newblock Segment anything.
\newblock \emph{arXiv preprint arXiv:2304.02643}.

\bibitem[{Lele et~al.(2019)Lele, Ross, Zhiyao, and Chenliang}]{Lele_Ross_Zhiyao_Chenliang_2019}
Lele, C.; Ross, K.; Zhiyao, D.; and Chenliang, X. 2019.
\newblock Hierarchical Cross-Modal Talking Face Generation With Dynamic Pixel-Wise Loss.

\bibitem[{Nagrani et~al.(2017)Nagrani, Chung, Zisserman et~al.}]{nagrani2017voxceleb}
Nagrani, A.; Chung, J.~S.; Zisserman, A.; et~al. 2017.
\newblock Voxceleb: a large-scale speaker identification dataset.
\newblock \emph{arXiv preprint arXiv:1706.08612}.

\bibitem[{Naruniec et~al.(2020)Naruniec, Helminger, Schroers, and Weber}]{Naruniec_Helminger_Schroers_Weber_2020}
Naruniec, J.; Helminger, L.; Schroers, C.; and Weber, R. 2020.
\newblock High‐Resolution Neural Face Swapping for Visual Effects.
\newblock \emph{Computer Graphics Forum}, 173–184.

\bibitem[{Newell, Yang, and Deng(2016)}]{newell2016stacked}
Newell, A.; Yang, K.; and Deng, J. 2016.
\newblock Stacked hourglass networks for human pose estimation.
\newblock In \emph{Computer Vision--ECCV 2016: 14th European Conference, Amsterdam, The Netherlands, October 11-14, 2016, Proceedings, Part VIII 14}, 483--499. Springer.

\bibitem[{Ren et~al.(2020{\natexlab{a}})Ren, Chai, Tulyakov, Fang, Shen, and Yang}]{ren2020human}
Ren, J.; Chai, M.; Tulyakov, S.; Fang, C.; Shen, X.; and Yang, J. 2020{\natexlab{a}}.
\newblock Human motion transfer from poses in the wild.
\newblock In \emph{Computer Vision--ECCV 2020 Workshops: Glasgow, UK, August 23--28, 2020, Proceedings, Part III 16}, 262--279. Springer.

\bibitem[{Ren et~al.(2020{\natexlab{b}})Ren, Yu, Chen, Li, and Li}]{ren2020deep}
Ren, Y.; Yu, X.; Chen, J.; Li, T.~H.; and Li, G. 2020{\natexlab{b}}.
\newblock Deep image spatial transformation for person image generation.
\newblock In \emph{Proceedings of the IEEE/CVF Conference on Computer Vision and Pattern Recognition}, 7690--7699.

\bibitem[{Siarohin et~al.(2019{\natexlab{a}})Siarohin, Lathuili{\`e}re, Tulyakov, Ricci, and Sebe}]{siarohin2019animating}
Siarohin, A.; Lathuili{\`e}re, S.; Tulyakov, S.; Ricci, E.; and Sebe, N. 2019{\natexlab{a}}.
\newblock Animating arbitrary objects via deep motion transfer.
\newblock In \emph{Proceedings of the IEEE/CVF Conference on Computer Vision and Pattern Recognition}, 2377--2386.

\bibitem[{Siarohin et~al.(2019{\natexlab{b}})Siarohin, Lathuili{\`e}re, Tulyakov, Ricci, and Sebe}]{siarohin2019first}
Siarohin, A.; Lathuili{\`e}re, S.; Tulyakov, S.; Ricci, E.; and Sebe, N. 2019{\natexlab{b}}.
\newblock First order motion model for image animation.
\newblock \emph{Advances in Neural Information Processing Systems}, 32.

\bibitem[{Siarohin et~al.(2021)Siarohin, Woodford, Ren, Chai, and Tulyakov}]{siarohin2021motion}
Siarohin, A.; Woodford, O.~J.; Ren, J.; Chai, M.; and Tulyakov, S. 2021.
\newblock Motion representations for articulated animation.
\newblock In \emph{Proceedings of the IEEE/CVF Conference on Computer Vision and Pattern Recognition}, 13653--13662.

\bibitem[{Tulyakov et~al.(2018)Tulyakov, Liu, Yang, and Kautz}]{tulyakov2018mocogan}
Tulyakov, S.; Liu, M.-Y.; Yang, X.; and Kautz, J. 2018.
\newblock Mocogan: Decomposing motion and content for video generation.
\newblock In \emph{Proceedings of the IEEE Conference on Computer Vision and Pattern Recognition}, 1526--1535.

\bibitem[{Tumanyan et~al.(2022)Tumanyan, Bar-Tal, Bagon, and Dekel}]{tumanyan2022splicing}
Tumanyan, N.; Bar-Tal, O.; Bagon, S.; and Dekel, T. 2022.
\newblock Splicing vit features for semantic appearance transfer.
\newblock In \emph{Proceedings of the IEEE/CVF Conference on Computer Vision and Pattern Recognition}, 10748--10757.

\bibitem[{Wang et~al.(2018{\natexlab{a}})Wang, Liu, Zhu, Liu, Tao, Kautz, and Catanzaro}]{wang2018video}
Wang, T.-C.; Liu, M.-Y.; Zhu, J.-Y.; Liu, G.; Tao, A.; Kautz, J.; and Catanzaro, B. 2018{\natexlab{a}}.
\newblock Video-to-video synthesis.
\newblock \emph{arXiv preprint arXiv:1808.06601}.

\bibitem[{Wang et~al.(2018{\natexlab{b}})Wang, Alameda-Pineda, Xu, Fua, Ricci, and Sebe}]{wang2018every}
Wang, W.; Alameda-Pineda, X.; Xu, D.; Fua, P.; Ricci, E.; and Sebe, N. 2018{\natexlab{b}}.
\newblock Every smile is unique: Landmark-guided diverse smile generation.
\newblock In \emph{Proceedings of the IEEE Conference on Computer Vision and Pattern Recognition}, 7083--7092.

\bibitem[{Wang et~al.(2022)Wang, Yang, Bremond, and Dantcheva}]{wang2022latent}
Wang, Y.; Yang, D.; Bremond, F.; and Dantcheva, A. 2022.
\newblock Latent image animator: Learning to animate images via latent space navigation.
\newblock \emph{arXiv preprint arXiv:2203.09043}.

\bibitem[{Wiles et~al.(2018)Wiles, Koepke, Zisserman et~al.}]{wiles2018x2face}
Wiles, O.; Koepke, A.; Zisserman, A.; et~al. 2018.
\newblock X2face: A network for controlling face generation using images, audio, and pose codes.
\newblock In \emph{Proceedings of the European Conference on Computer Vision}, 670--686.

\bibitem[{Zhao and Zhang(2022)}]{zhao2022thin}
Zhao, J.; and Zhang, H. 2022.
\newblock Thin-plate spline motion model for image animation.
\newblock In \emph{Proceedings of the IEEE/CVF Conference on Computer Vision and Pattern Recognition}, 3657--3666.

\bibitem[{Zhou et~al.(2021)Zhou, Sun, Wu, Loy, Wang, and Liu}]{Zhou_Sun_Wu_Loy_Wang_Liu_2021}
Zhou, H.; Sun, Y.; Wu, W.; Loy, C.~C.; Wang, X.; and Liu, Z. 2021.
\newblock Pose-Controllable Talking Face Generation by Implicitly Modularized Audio-Visual Representation.
\newblock In \emph{2021 IEEE/CVF Conference on Computer Vision and Pattern Recognition}.

\end{thebibliography}

\end{document}